\documentclass[11pt]{article}

\usepackage{acl}

\usepackage{times}
\usepackage{latexsym}
\usepackage[T1]{fontenc}
\usepackage[utf8]{inputenc}
\usepackage{microtype}
\usepackage{inconsolata}

\usepackage{graphicx}
\usepackage{booktabs}
\usepackage{multirow}
\usepackage{tabularx}
\usepackage{subcaption}
\usepackage{float}

\usepackage{amsmath}
\usepackage{amssymb}
\usepackage{amsfonts}

\usepackage{algorithm}
\usepackage{algorithmic}

\usepackage{url}
\usepackage{hyperref}
\usepackage{xcolor}

\newcommand{\loss}[1]{\mathcal{L}_{\text{#1}}}
\newcommand{\rouge}{\textsc{Rouge}}
\newcommand{\bleu}{\textsc{Bleu}}
\newcommand{\ewad}{\textsc{Ewad}}
\newcommand{\cpdp}{\textsc{Cpdp}}

\title{Reliability-Gated Multi-Teacher Distillation for Low-Resource Abstractive Summarization}

\author{
Dipto Sumit, Ankan Kumar Roy, Sadia Khair Rodela, Atia Haque Asha, Mourchona Afrin \\
Niloy Farhan, Farig Yousuf Sadeque \\
BRAC University
}

\begin{document}
\maketitle

\begin{abstract}
We study multi-teacher knowledge distillation for low-resource abstractive summarization from a \emph{reliability-aware} perspective. We introduce \textbf{EWAD} (Entropy-Weighted Agreement-Aware Distillation), a token-level mechanism that routes supervision between teacher distillation and gold supervision based on inter-teacher agreement, and \textbf{CPDP} (Capacity-Proportional Divergence Preservation), a geometric constraint on the student’s position relative to heterogeneous teachers. Across two Bangla datasets, 13 BanglaT5 ablations, and eight Qwen-2.5 experiments, we find that \emph{logit-level KD provides the most reliable gains}, while more complex distillation improves semantic similarity for short summaries but degrades longer outputs. Cross-lingual pseudo-label KD across ten languages retains 71--122\% of teacher ROUGE-L at 3.2$\times$ compression. A human-validated multi-judge LLM evaluation further reveals calibration bias in single-judge pipelines. Overall, our results show that reliability-aware distillation helps characterize when multi-teacher supervision improves summarization and when data scaling outweighs loss engineering.
\end{abstract}

% ===================================================================
%  §1  INTRODUCTION
% ===================================================================
\section{Introduction}
\label{sec:intro}

Large sequence-to-sequence models have greatly improved abstractive summarization~\cite{lewis-etal-2020-bart,raffel2020exploring,xue-etal-2021-mt5}, but their computational cost limits deployment in low-resource settings. Knowledge distillation (KD)~\cite{hinton2015distilling} addresses this by transferring knowledge from a large \emph{teacher} to a smaller \emph{student}. However, distillation for generative tasks is fragile: teacher predictions vary across tokens, and disagreement between teachers can introduce noisy supervision.

Most prior work relies on single-teacher KD or static aggregation of multiple teachers~\cite{kim-rush-2016-sequence,you2017learning,fukuda2017efficient}, assuming teacher predictions can be safely averaged. In practice, teachers often disagree on entities and phrasing, suggesting that supervision should be applied selectively.

We therefore study \emph{reliability-aware multi-teacher distillation}. We introduce \textbf{EWAD} (Entropy-Weighted Agreement-Aware Distillation), a token-level objective that routes supervision between teacher KD and gold supervision based on teacher confidence and inter-teacher agreement, and \textbf{CPDP} (Capacity-Proportional Divergence Preservation), a geometric constraint regulating the student’s divergence relative to heterogeneous teachers. Experiments on two Bangla summarization datasets show that \emph{logit-level KD provides the most consistent gains}, while additional KD components improve semantic similarity for short summaries but degrade longer outputs. In Qwen-2.5 experiments, reliability-aware routing does not outperform direct fine-tuning, suggesting a capacity ceiling when student quality approaches the teacher. Cross-lingual pseudo-label KD transfers across ten languages with 71--122\% teacher ROUGE-L retention. Rather than treating reliability-aware distillation purely as a performance optimization, we use it as a framework to study when multi-teacher supervision improves generation and when it introduces noise.

\noindent\textbf{Our contributions are:}
\begin{itemize}\itemsep-1pt
\item We introduce \textbf{EWAD}, a reliability-gated distillation objective that dynamically switches between teacher supervision and gold supervision based on inter-teacher agreement.
\item We introduce \textbf{CPDP}, a capacity-aware geometric regularizer constraining the student’s divergence relative to heterogeneous teachers.
\item Through extensive experiments, we show that logit-level KD dominates more complex KD objectives and that multi-component KD is strongly output-length dependent.
\item We demonstrate cross-lingual pseudo-label distillation across ten languages and highlight calibration bias in single-judge LLM evaluation.
\end{itemize}

% ===================================================================
%  §2  RELATED WORK
% ===================================================================
\section{Related Work}
\label{sec:related}

\paragraph{KD for sequence generation.}
\citet{hinton2015distilling} introduced distillation via softened logits; \citet{kim-rush-2016-sequence} extended it to sequence-level KD.
Multi-teacher KD~\cite{you2017learning,fukuda2017efficient} and teacher assistants~\cite{mirzadeh2020improved} address capacity gaps, while \citet{park2019relational} transfers inter-sample relationships.
All prior methods use \emph{fixed or globally learned} weights; to our knowledge, prior work does not explicitly gate distillation per token using inter-teacher agreement nor enforce capacity-proportional divergence.

\paragraph{Low-resource and multilingual summarization.}
XL-Sum~\cite{hasan-etal-2021-xl} covers 44 languages; BanglaT5~\cite{bhattacharjee-etal-2023-banglanlg} is the first Bangla seq2seq model.
Cross-lingual transfer~\cite{chi-etal-2021-mt6} and adapters~\cite{pfeiffer-etal-2020-adapterfusion} reduce adaptation cost but not inference cost.
We show pseudo-label KD retains 71--122\% \rouge{}-L across ten languages at 3.2$\times$ compression.

\paragraph{Adaptive and token-level signals.}
Focal loss~\cite{lin2017focal}, TinyBERT~\cite{jiao-etal-2020-tinybert}, and token-level weighting~\cite{wen2023f} move beyond uniform losses.
\ewad{} uniquely decomposes gating into \emph{confidence} and \emph{agreement}; \cpdp{} adds geometric constraints absent from prior adaptive KD.

% === Remaining sections — to be written incrementally ===

% ===================================================================
%  §3  METHODOLOGY
% ===================================================================
\section{Methodology}
\label{sec:method}

Our framework combines a \emph{heterogeneous teacher ensemble} (same-vocabulary logit KD and cross-architecture pseudo-label supervision) with two reliability-aware training components: \ewad{}, which dynamically routes supervision between teacher KD and gold supervision based on inter-teacher agreement, and \cpdp{}, which imposes a capacity-aware geometric constraint on the student's divergence relative to heterogeneous teachers.

% ------------------------------------------------------------------
% Figures
% ------------------------------------------------------------------
\begin{figure*}[t]
\centering
\includegraphics[width=\textwidth]{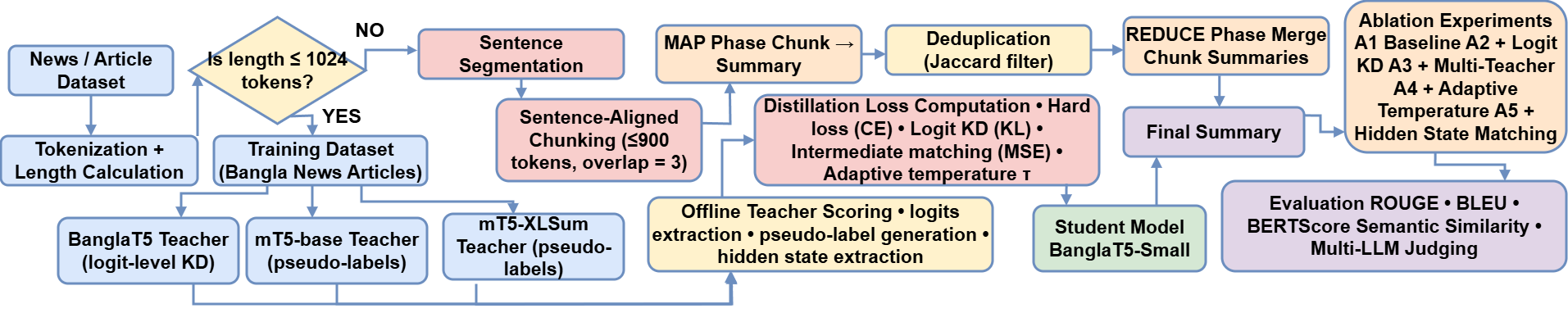}
\caption{End-to-end framework. Documents are length-routed to the multi-teacher KD branch or MapReduce module. Three teachers provide logit and pseudo-label supervision across five ablation stages.}
\label{fig:pipeline}
\end{figure*}

\begin{figure}[t]
\centering
\includegraphics[width=\columnwidth]{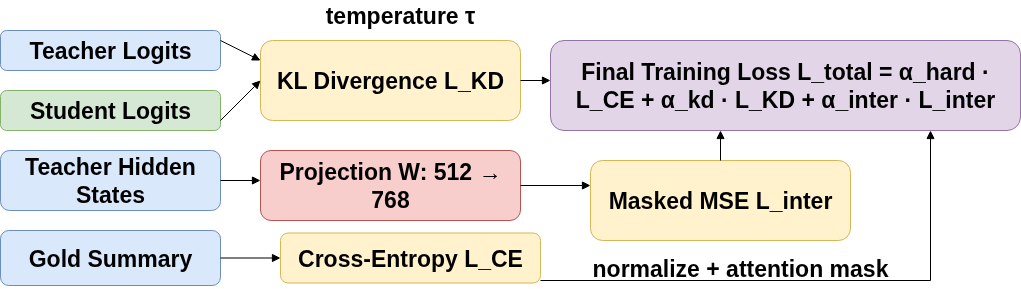}
\caption{Standard distillation loss (Eq.~\ref{eq:base-loss}): $\loss{KD}$ (softened KL), $\loss{inter}$ (projected MSE), and $\loss{CE}$ (gold cross-entropy).}
\label{fig:loss}
\end{figure}

% ------------------------------------------------------------------
\subsection{Framework Overview}
\label{ssec:overview}
% ------------------------------------------------------------------

Given document $D = (x_1, \ldots, x_n)$, the goal is to generate summary $S = (y_1, \ldots, y_m)$ maximizing $P(S \mid D)$.
Since transformers are bounded by context window $L{=}1024$, the pipeline begins with length-aware routing:
\begin{equation}
\label{eq:routing}
R(D) = \begin{cases}
  \text{DirectKD}(D) & \text{if } n \leq L \\
  \text{MapReduce}(D) & \text{if } n > L
\end{cases}
\end{equation}
Short documents enter the KD branch; longer ones are routed to MapReduce (\S\ref{ssec:mapreduce}).
Teachers are scored \emph{offline} and cached; the student trains under a composite loss (CE + KL + optional MSE on hidden states).
A five-stage ablation (A1--A5) and a separate eight-experiment \ewad{}+\cpdp{} ablation isolate each component's contribution.

% ------------------------------------------------------------------
\subsection{Multi-Teacher Knowledge Transfer}
\label{ssec:multi-teacher}
% ------------------------------------------------------------------

\paragraph{Same-Vocabulary Logit-Level KD.}
When teacher and student share the same vocabulary as with BanglaT5 (247.6M) $\to$ BanglaT5-small (109.9M), or within the Qwen-2.5 family (32B/14B $\to$ 3B) we perform \emph{logit-level} distillation.
The teacher is run in teacher-forced mode, producing $p_T^t$ at each step; the student minimizes KL$(p_T^t \| p_S^t)$ softened by $\tau$ (\S\ref{ssec:temperature}).
For Qwen-2.5, teachers use NF4 quantization and only top-$k$ ($k{=}50$) log-probs are cached per position.

\paragraph{Cross-Architecture Pseudo-Label Transfer.}
When tokenizers are incompatible (mT5 250K vs.\ BanglaT5 32K), we use sequence-level distillation~\cite{kim-rush-2016-sequence}: each teacher generates pseudo-summaries (beam $B{=}4$), stored as text and re-tokenized at training time.
Pseudo-labels replace gold summaries with probability $p_{\text{pseudo}}{=}0.3$ in A3--A5, using two mT5 teachers ($\sim$580M each), creating a three-teacher heterogeneous ensemble.

% ------------------------------------------------------------------
\subsection{Loss Functions}
\label{ssec:losses}
% ------------------------------------------------------------------

A fundamental tension in multi-teacher distillation for abstractive summarization is that teachers are not equally reliable at every generation step. Standard KD losses treat every token and every teacher identically, providing no mechanism for the student to distinguish confident consensus from noisy disagreement.
We address this through a three-tier loss design: (i)~a \emph{standard distillation baseline} that establishes gold-anchored supervision (\S\ref{sssec:standard-loss}), (ii)~\ewad{}, which introduces token-level reliability gating so the student selectively trusts teachers only when they are both confident and consistent (\S\ref{sssec:ewad}), and (iii)~\cpdp{}, which constrains the student's position in distribution space relative to each teacher, enforcing a geometric coherence absent from all prior KD objectives (\S\ref{sssec:cpdp}).

\subsubsection{Standard Distillation Loss}
\label{sssec:standard-loss}

Before introducing our novel components, we establish a baseline objective that keeps gold supervision dominant while transferring teacher knowledge critical for maintaining summary quality when compressing models for deployment on resource-constrained devices.
The base training objective for the BanglaT5 experiments (Figure~\ref{fig:loss}) combines three loss terms:
\begin{equation}
\label{eq:base-loss}
\loss{total} = \alpha_{\text{hard}} \cdot \loss{CE} + \alpha_{\text{kd}} \cdot \loss{KD} + \alpha_{\text{inter}} \cdot \loss{inter}
\end{equation}
where $\alpha_{\text{hard}} = 1 - \alpha_{\text{kd}} - \alpha_{\text{inter}}$ so the weights sum to one.

\paragraph{Cross-Entropy Loss ($\loss{CE}$).} Gold-label supervision anchors the student to human-written summaries, preventing the distribution drift that is particularly dangerous in low-resource settings where training data is scarce and every reference signal is valuable. Standard token-level negative log-likelihood on the gold summary:
\begin{equation}
\loss{CE} = -\frac{1}{T}\sum_{t=1}^{T} \log p_S^t(y_t^*)
\end{equation}
where $y_t^*$ is the gold token at position $t$ and $T$ is the summary length.

\paragraph{Logit-Level KD Loss ($\loss{KD}$).} While hard labels encode only the argmax token, teacher logits carry a full distribution over the vocabulary capturing semantic proximity between candidate tokens (e.g., synonyms receiving similar probability mass) that is especially informative for abstractive generation, where multiple valid paraphrases exist. KL divergence between temperature-softened teacher and student distributions:
\begin{equation}
\loss{KD} = \tau^2 \cdot \text{KL}\!\left(\,\text{softmax}\!\left(\frac{\mathbf{z}_T}{\tau}\right) \;\Big\|\; \text{softmax}\!\left(\frac{\mathbf{z}_S}{\tau}\right)\right)
\end{equation}
where $\mathbf{z}_T$ and $\mathbf{z}_S$ are teacher and student logits, respectively, and $\tau$ is the distillation temperature (fixed at 0.8 in A2--A3; adaptive in A4--A5).
The $\tau^2$ scaling maintains gradient magnitude parity with the cross-entropy term~\cite{hinton2015distilling}.

\paragraph{Intermediate Matching Loss ($\loss{inter}$).}
Logit-level and label-level losses supervise only the output layer; yet for summarization, the encoder must learn to identify salient content, resolve coreference, and compress discourse structure capacities encoded in intermediate representations.
By aligning encoder hidden states, we transfer these structural competencies directly, stabilizing learning especially under aggressive compression. We align encoder hidden states via a learned projection $\mathbf{W} \in \mathbb{R}^{512 \times 768}$ (student $d{=}512$ $\to$ teacher $d{=}768$):
\begin{equation}
\loss{inter} = \frac{1}{|\mathcal{M}|}\sum_{t \in \mathcal{M}} \left\| \overline{\mathbf{h}}_S^t - \overline{\mathbf{h}}_T^t \right\|^2, \quad \overline{\mathbf{h}} = \frac{\mathbf{h}}{\|\mathbf{h}\|_2}
\end{equation}
where $\mathcal{M}$ is the set of non-padding positions and $\overline{\mathbf{h}}_S^t = \overline{\mathbf{W}\,\mathbf{h}_S^t}$.
Active only in A5 ($\alpha_{\text{inter}}{=}0.1$).

% --- EWAD ---

\subsubsection{\ewad{}: Entropy-Weighted Agreement-Aware Distillation}
\label{sssec:ewad}

\begin{figure}[h]
\centering
\includegraphics[width=\columnwidth]{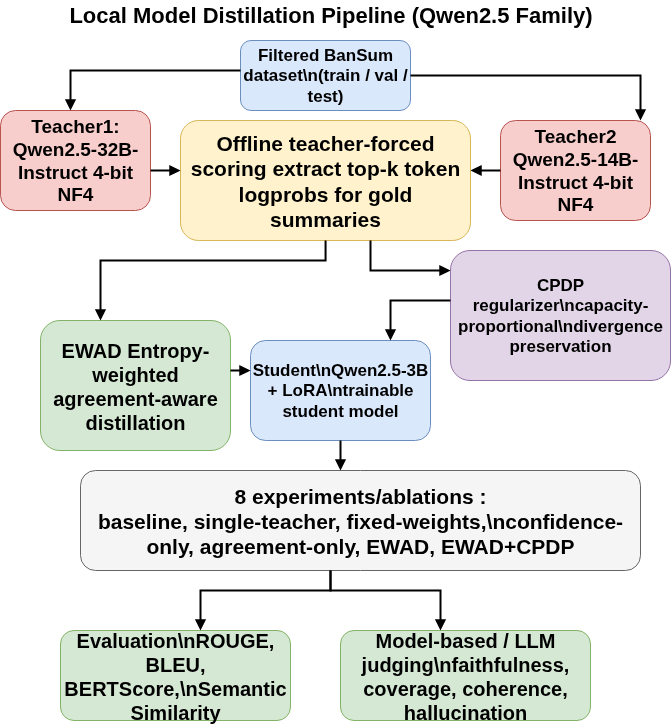}
\caption{Dual-teacher \ewad{}+\cpdp{} with Qwen-2.5 (32B + 14B $\to$ 3B + LoRA). Eight ablation experiments isolate each component.}
\label{fig:dual-teacher}
\end{figure}

The standard loss above treats every teacher signal as equally trustworthy at every token position an assumption that breaks down in multi-teacher summarization.
Consider a position where one teacher confidently predicts a factual entity while the other spreads probability across unrelated tokens: blindly averaging their logits injects noise that, in autoregressive generation, propagates through all subsequent tokens.
\ewad{} addresses this with a two-axis reliability decomposition unique to our framework:
\emph{confidence} (which teacher to trust, based on entropy) and \emph{agreement} (whether to trust teachers \emph{at all}, based on distributional divergence).
When both teachers are confident and agree, the student receives rich soft-label supervision; when they conflict, the loss automatically falls back to gold labels, preventing the student from learning from contradictory signals.
This token-level gating enables reliability-aware distillation by dynamically routing supervision between teacher KD and gold supervision based on inter-teacher agreement.
We describe the four-step computation below for teachers $T_1$, $T_2$ and student $S$ (Figure~\ref{fig:dual-teacher}).

\paragraph{Step 1: Teacher Confidence.}
The first axis measures how decisive each teacher is at each generation step.
A teacher that concentrates probability on a few tokens carries a stronger, more informative signal than one with a flat distribution a distinction especially important in summarization, where content selection requires the model to commit to specific entities and phrasings.
At each summary token position $t$, we compute each teacher's confidence as the complement of its normalized entropy:
\begin{equation}
\label{eq:confidence}
C_i^t = 1 - \frac{H(p_i^t)}{\log |V|}
\end{equation}
where $H(p_i^t) = -\sum_{v} p_i^t(v)\log p_i^t(v)$ and $|V|$ is the vocabulary size.
$C_i^t \in [0,1]$; values near 1 indicate the teacher concentrates probability mass on a few tokens.

\paragraph{Step 2: Confidence-Proportional Weights.}
Given per-token confidence scores, we must convert them into a weighting scheme that lets the more certain teacher dominate without hard-switching, which would discard useful information from the less confident teacher. Each teacher's influence is determined by a softmax over their confidence scores:
\begin{align}
\label{eq:weights}
w_{T_1}^t &= \frac{\exp\!\bigl(C_{T_1}^t / \tau_w\bigr)}{\exp\!\bigl(C_{T_1}^t / \tau_w\bigr) + \exp\!\bigl(C_{T_2}^t / \tau_w\bigr)},
\nonumber\\
w_{T_2}^t &= 1 - w_{T_1}^t
\end{align}
with $\tau_w{=}1.0$.

\paragraph{Step 3: Agreement Gate.}
Confidence alone is insufficient: two teachers can each be highly confident yet predict entirely different tokens a failure mode where naive confidence weighting would still inject harmful supervision.
The second axis of \ewad{} measures inter-teacher \emph{agreement} via Jensen--Shannon divergence (JSD), a symmetric and bounded divergence that naturally captures distributional overlap:
\begin{equation}
\label{eq:agreement}
A_t = 1 - \frac{\text{JSD}(p_{T_1}^t \| p_{T_2}^t)}{\log 2}
\end{equation}
where $M = \tfrac{1}{2}(p_{T_1}^t + p_{T_2}^t)$ is the implicit midpoint distribution in the Jensen-Shannon divergence: $\text{JSD}(p \| q) = \frac{1}{2}\text{KL}(p \| M) + \frac{1}{2}\text{KL}(q \| M)$.
The score is passed through a sigmoid gate that creates a smooth but decisive transition between trust and distrust:
\begin{equation}
\label{eq:gate}
\lambda_t = \sigma\!\big(k \cdot (A_t - \delta)\big)
\end{equation}
with $k{=}5.0$ and $\delta{=}0.5$; $\lambda_t{\approx}1$ when teachers agree, $\lambda_t{\approx}0$ otherwise.

\paragraph{Step 4: EWAD Loss.}
The agreement gate enables the core novelty of \ewad{}: a \emph{dynamic routing} between two supervision regimes.
When teachers agree strongly ($\lambda_t \approx 1$), the student receives confidence-weighted soft-label KD that transfers rich distributional knowledge; when they conflict ($\lambda_t \approx 0$), the loss falls back to gold-label cross-entropy, treating the position as if no teacher supervision were available rather than polluting the student with contradictory signals. The weighted KD loss and gold-label CE loss are blended via the agreement gate:
\begin{equation}
\label{eq:ewad}
\loss{EWAD} = \frac{1}{T}\sum_{t=1}^{T}\bigl[\lambda_t\,\loss{KD}^t + (1{-}\lambda_t)\,\loss{CE}^t\bigr]
\end{equation}
where the per-token distillation and gold-label terms are
\begin{align}
\loss{KD}^t &= w_{T_1}^t \cdot \text{KL}(p_{T_1}^t \| p_S^t) \nonumber\\
            &\quad + w_{T_2}^t \cdot \text{KL}(p_{T_2}^t \| p_S^t), \label{eq:ewad-kd}\\
\loss{CE}^t &= -\log p_S^t(y_t^*). \label{eq:ewad-ce}
\end{align}

% --- CPDP ---

\subsubsection{\cpdp{}: Capacity-Proportional Divergence Preservation}
\label{sssec:cpdp}
While EWAD determines when teacher supervision should be trusted, CPDP constrains where the student should lie in distribution space relative to teachers of different capacities.
\ewad{} governs \emph{when} to trust teachers; \cpdp{} addresses a complementary question: \emph{where should the student sit in distribution space relative to its teachers?}
In a multi-teacher setup with heterogeneous capacities (e.g., 32B and 14B distilling into 3B), the student should not be equidistant from both teachers it should be closer to the smaller teacher whose representational capacity it more closely matches, and further from the larger teacher.
Moreover, the gap between the student's divergences to each teacher should reflect the teachers' own mutual divergence.
Without this constraint, the student can collapse toward one teacher while ignoring the other, or occupy an arbitrary region of distribution space that is geometrically inconsistent with the capacity hierarchy.

\cpdp{} enforces this structure by penalizing deviations from a capacity-proportional divergence ratio.
Let the two teachers have distributions $p_{T_1}$ (larger) and $p_{T_2}$ (smaller), and the student $p_S$.
We define the target divergence gap as the inter-teacher divergence $\Delta^* = \text{KL}(p_{T_1} \| p_{T_2})$, and regularize the student's \emph{normalized} divergence difference to match it:
\begin{equation}
\label{eq:cpdp}
\loss{CPDP} = \left(\frac{\text{KL}(p_{T_1} \| p_S)}{H(p_S)} - \frac{\text{KL}(p_{T_2} \| p_S)}{H(p_S)} - \Delta^*\right)^{\!2}
\end{equation}
The normalization by student entropy $H(p_S)$ converts raw KL values into capacity-relative units, ensuring the constraint is meaningful regardless of the student's absolute entropy level.
Critically, $H(p_S)$ is \textbf{detached} during backpropagation to prevent the trivial solution of minimizing the loss by inflating student entropy.
The inter-teacher divergence $\Delta^*$ is computed once and held fixed throughout training, serving as a geometric anchor.
Weighted by $\mu{=}0.05$; per-token values clamped at 100.0 for numerical stability. This regularizer is, to our knowledge, the first to impose a \emph{capacity-aware geometric constraint} on the student's position in a multi-teacher KD framework moving beyond scalar loss weighting to enforce structural relationships in distribution space.

% --- Combined ---

\subsubsection{Combined Training Objective}
\label{sssec:combined}

\ewad{} and \cpdp{} address orthogonal failure modes: \ewad{} prevents the student from learning from \emph{unreliable} teacher signals (a temporal problem which tokens to trust), while \cpdp{} prevents the student from occupying a \emph{geometrically incoherent} position in distribution space (a structural problem where to sit relative to teachers).
Their combination yields a training objective that is both reliability-gated and capacity-aware:
\begin{equation}
\label{eq:total}
\loss{total} = \loss{EWAD} + \mu \cdot \loss{CPDP}
\end{equation}
BanglaT5 ablations use Eq.~\ref{eq:base-loss} with $\alpha_{\text{kd}}{=}0.01$, $\alpha_{\text{inter}} \in \{0, 0.1\}$.

% ------------------------------------------------------------------
\subsection{Confidence-Adaptive Temperature Scaling}
\label{ssec:temperature}
% ------------------------------------------------------------------

In A4--A5, we adapt $\tau$ per sample so confident teachers provide sharper supervision while uncertain teachers are softened, allowing the model to self-adjust distillation strength based on input difficulty. The temperature is adapted \emph{per sample} based on the teacher's mean token-level entropy:
\begin{align}
\label{eq:adaptive-tau-H}
\bar{H} &= \frac{1}{|\mathcal{M}|}\sum_{t \in \mathcal{M}} H(p_T^t), \\
\label{eq:adaptive-tau}
\tau &= \tau_{\min} + (\tau_{\max} - \tau_{\min}) \cdot \sigma\!\bigl(\bar{H} - \bar{H}_{\text{batch}}\bigr)
\end{align}
where $\tau_{\min}{=}0.5$, $\tau_{\max}{=}2.0$, $\bar{H}_{\text{batch}}$ is the batch mean entropy, and $\sigma$ is the sigmoid function.
Low-entropy samples get sharper supervision ($\tau{\to}0.5$); high-entropy samples get smoother distributions ($\tau{\to}2.0$).

% ------------------------------------------------------------------
\subsection{Long-Document Handling via MapReduce}
\label{ssec:mapreduce}
% ------------------------------------------------------------------

Documents exceeding the 1\,024-token context window (the ``NO'' branch in Figure1 ~\ref{fig:pipeline}) are processed by a two-stage MapReduce pipeline rather than being truncated.

\paragraph{Sentence-Aligned Chunking.}
Sentences $D = (s_1, \ldots, s_k)$ are accumulated into chunks up to $C{=}900$ tokens:
\begin{equation}
c_i = (s_{a_i}, \ldots, s_{b_i}), \quad |c_i| \leq C
\end{equation}
Consecutive chunks overlap by $o{=}3$ sentences:
\begin{equation}
c_{i+1} = (s_{b_i - o + 1}, \ldots, s_{b_{i+1}})
\end{equation}

\paragraph{MAP Phase.}
Each chunk is independently summarized by a BanglaT5 MAP model fine-tuned on full-document inputs: $\hat{s}_i = f_\theta(c_i)$.

\paragraph{Deduplication.}
Because overlapping chunks may produce redundant content, we remove duplicate sentences across chunk summaries using Jaccard similarity:
\begin{equation}
J(A, B) = \frac{|A \cap B|}{|A \cup B|}
\end{equation}
Sentence pairs with $J > 0.75$ are deduplicated.

\paragraph{REDUCE Phase.}
Deduplicated summaries are concatenated and passed to a REDUCE model (transfer-learned from MAP): $S = f_\phi(\text{concat}(\hat{S}_{\text{map}}))$, applied recursively.

% ------------------------------------------------------------------
\subsection{Cross-Lingual Extension}
\label{ssec:cross-lingual}
% ------------------------------------------------------------------

We apply offline pseudo-label KD to ten languages (Hindi, Urdu, Persian, Amharic, Hausa, Nepali, Pashto, Indonesian, Turkish, Swahili): a fine-tuned mT5-XLSum teacher (966M) generates pseudo-summaries, and mT5-small (300M, 3.2$\times$ compression) trains on them.
Shared tokenizers eliminate vocabulary mismatch; offline caching makes training 100$\times$ faster than online KD.
Notably, the entire framework including \ewad{}, \cpdp{}, and the MapReduce pipeline is \emph{language-agnostic}: it relies on no language-specific heuristics beyond tokenization, and the cross-lingual results confirm that it transfers without modification across typologically diverse languages and scripts.

% ------------------------------------------------------------------
% Figures
% ------------------------------------------------------------------
% \begin{figure*}[t]
% \centering
% \includegraphics[width=\textwidth]{figure1_unified_pipeline.drawio.png}
% \caption{End-to-end framework. Documents are length-routed to the multi-teacher KD branch or MapReduce module. Three teachers provide logit and pseudo-label supervision across five ablation stages.}
% \label{fig:pipeline}
% \end{figure*}

% \begin{figure}[t]
% \centering
% \includegraphics[width=\columnwidth]{figure3_loss_design.drawio.png}
% \caption{Standard distillation loss (Eq.~\ref{eq:base-loss}): $\loss{KD}$ (softened KL), $\loss{inter}$ (projected MSE), and $\loss{CE}$ (gold cross-entropy).}
% \label{fig:loss}
% \end{figure}

% \begin{figure}[t]
% \centering
% \includegraphics[width=\columnwidth]{dual_teacher_kd.drawio.png}
% \caption{Dual-teacher \ewad{}+\cpdp{} with Qwen-2.5 (32B + 14B $\to$ 3B + LoRA). Eight ablation experiments isolate each component.}
% \label{fig:dual-teacher}
% \end{figure}

% ===================================================================
%  §4  EXPERIMENTAL SETUP
% ===================================================================
\section{Experimental Setup : Datasets, Models, Metrics, and Compute}
\label{sec:setup}

% \subsection{}
% \label{ssec:datasets}
% We use three datasets: BTS (68K Bangla news article–headline pairs, $\sim$8-token targets), BanSum (141K document–summary pairs, $\sim$106–248-token targets; a 20K filtered subset used for EWAD+CPDP), and XL-Sum subsets covering ten languages (≤512 source tokens). Table~\ref{tab:models} lists all models. For BanglaT5 experiments, the teacher shares the student vocabulary while two mT5 models generate pseudo-labels. For EWAD+CPDP, Qwen-2.5 teachers (NF4) distill into Qwen-2.5-3B using LoRA ($r=64$, $\alpha=128$). We report ROUGE-1/2/L \citep{lin2004rouge}, BLEU \citep{papineni2002bleu}, BERTScore F1 \citep{zhang2020bertscore}, semantic similarity, and LLM-judge scores. All experiments were conducted on consumer GPUs (RTX 5090/5080/4070 Ti Super), totaling approximately 320 GPU-hours.

We use three datasets: BTS (68K Bangla news article–headline pairs, $\sim$8-token targets) \citep{hasanmoni_bengali_summarization}, BanSum (141K document–summary pairs, $\sim$106–248-token targets; a 20K filtered subset used for EWAD+CPDP) \citep{hasan2024bansum}, and XL-Sum subsets covering ten languages ($\leq$512 source tokens) \citep{hasan-etal-2021-xl}. Table~\ref{tab:models} lists all models. For BanglaT5 experiments, the teacher shares the student vocabulary while two mT5 models generate pseudo-labels. For EWAD+CPDP, Qwen-2.5 teachers (NF4) distill into Qwen-2.5-3B using LoRA ($r=64$, $\alpha=128$). We report ROUGE-1/2/L \citep{lin2004rouge}, BLEU \citep{papineni2002bleu}, BERTScore F1 \citep{zhang2020bertscore}, semantic similarity, and LLM-judge scores. All experiments were conducted on consumer GPUs (RTX 5090/5080/4070 Ti Super), totaling approximately 320 GPU-hours.

% \textbf{BTS} (68K bangla news article--headline pairs, $\sim$8-token targets), \textbf{BanSum} (141K document--summary pairs, $\sim$106--248-token targets; 20K filtered subset for \ewad{}+\cpdp{}), and \textbf{XL-Sum} subsets for ten languages ($\leq$512 source tokens).
% Table~\ref{tab:models} lists all models.
% For BanglaT5, the teacher shares the student vocabulary; two mT5 models supply pseudo-labels.
% For \ewad{}+\cpdp{}, Qwen-2.5 teachers (NF4) distill into Qwen-2.5-3B via LoRA ($r{=}64$, $\alpha{=}128$).

\begin{table}[h]
\centering\footnotesize
\setlength{\tabcolsep}{4pt}
\begin{tabular}{@{}llr@{}}
\toprule
\textbf{Role} & \textbf{Model} & \textbf{Params} \\
\midrule
\multicolumn{3}{@{}l}{\textit{BanglaT5 Ablation (A1--A5)}} \\
Teacher (logit) & BanglaT5 & 247.6M \\
Teacher (pseudo) & mT5-base / mT5-XLSum & ${\sim}$580M \\
Student & BanglaT5-small & 109.9M \\
\addlinespace[3pt]
\multicolumn{3}{@{}l}{\textit{\ewad{}+\cpdp{} (Qwen-2.5)}} \\
Teachers & Qwen2.5-32B / 14B-Inst. & 32B / 14B \\
Student & Qwen2.5-3B + LoRA & 3B \\
\addlinespace[3pt]
\multicolumn{3}{@{}l}{\textit{Cross-Lingual}} \\
Teacher / Student & mT5-XLSum / mT5-small & 966M / 300M \\
\bottomrule
\end{tabular}
\caption{Teacher and student model configurations.}
\label{tab:models}
\end{table}

% \subsection{Training and Ablation}
% \label{ssec:training}

% All runs use AdamW with bf16 mixed precision and gradient clipping at 1.0.
% \textbf{BanglaT5}: batch 32, lr $5{\times}10^{-5}$, 5 epochs, early stopping on \rouge{}-L.
% \textbf{Qwen-2.5}: batch 32, lr $2{\times}10^{-4}$, cosine annealing, 2 epochs.
% \textbf{Cross-lingual}: BanglaT5 template on offline pseudo-labels.
% BanglaT5 ablation: \textbf{A1}~(CE), \textbf{A2}~(+logit KD), \textbf{A3}~(+pseudo-labels), \textbf{A4}~(+adaptive~$\tau$), \textbf{A5}~(+intermediate matching).
% Qwen-2.5: 8 experiments isolating each component on the 20K subset; best scaled to full 141K.

% \subsection{Metrics and Compute}
% \label{ssec:metrics}

% We report \rouge{}-1/2/L~\cite{lin2004rouge}, \bleu{}~\cite{papineni2002bleu}, BERTScore F1~\cite{zhang2020bertscore}, semantic similarity, and LLM-judge scores.
% All experiments ran on consumer-grade GPUs (RTX~5090/5080/4070\,Ti\,Super); total compute $\sim$320 GPU-hours.

% ===================================================================
%  §5  RESULTS
% ===================================================================
\section{Results}
\label{sec:results}

% ------------------------------------------------------------------
\subsection{BanglaT5 Ablation: Bengali Text Summarization}
\label{ssec:res-bts}
% ------------------------------------------------------------------

Table~\ref{tab:bts-rouge} presents the five-stage ablation on the BTS dataset (8,033 test samples).
A2 (single-teacher logit KD) achieves the highest \rouge{} scores, retaining 93.6\% of teacher \rouge{}-L.
Adding further components (A3--A5) slightly reduces n-gram overlap; however, semantic similarity increases monotonically from A1 to A5 ($0.868 \to 0.870$), indicating that each component injects meaningful representational knowledge even when surface overlap saturates.
A5 achieves the highest semantic similarity (0.8695), showing that encoder matching transfers structure beyond ROUGE.

\begin{table}[h]
\centering\small
\setlength{\tabcolsep}{2.7pt}
\begin{tabular}{lcccccc}
\toprule
\textbf{Config} & \textbf{R-1} & \textbf{R-2} & \textbf{R-L} & \textbf{BLEU} & \textbf{BS\,F1} & \textbf{Sem} \\
\midrule
A1 Baseline   & .3943 & .2321 & .3794 & 15.79 & .7863 & .8678 \\
A2 +LogitKD   & \textbf{.3945} & \textbf{.2321} & \textbf{.3797} & 15.49 & .7867 & .8683 \\
A3 +Pseudo    & .3924 & .2301 & .3777 & 15.20 & .7855 & .8689 \\
A4 +Adapt.\,$\tau$ & .3914 & .2296 & .3767 & 15.02 & .7846 & .8693 \\
A5 +InterMatch & .3914 & .2299 & .3769 & 15.08 & .7848 & \textbf{.8695} \\
\midrule
\textit{Teacher} & \textit{.4234} & \textit{.2512} & \textit{.4058} & \textit{16.47} & \textit{.7917} & \textit{.8722} \\
\bottomrule
\end{tabular}
\caption{BTS ablation results (BanglaT5-small student, 109.9M). R = \rouge{}, BS = BERTScore, Sem = Semantic Similarity. Best student scores in \textbf{bold}.}
\label{tab:bts-rouge}
\end{table}
% ------------------------------------------------------------------
\subsection{BanglaT5 Ablation: BanSum}
\label{ssec:res-bansum}
% ------------------------------------------------------------------

Table~\ref{tab:bansum-rouge} shows results on the BanSum dataset (14,120 test samples).
A2 dominates across \emph{all} metrics \rouge{}, \bleu{}, BERTScore, and semantic similarity achieving 95.9\% of teacher \rouge{}-L.
Unlike BTS, additional components (A3--A5) consistently degrade performance.
The contrasting behavior is explained by BanSum's substantially longer outputs ($\sim$106--248 tokens vs.\ $\sim$8 tokens in BTS): the extra regularization from pseudo-labels and adaptive temperature introduces harmful noise on longer, more diverse summaries.

\begin{table}[h]
\centering\small
\setlength{\tabcolsep}{2.7pt}
\begin{tabular}{lcccccc}
\toprule
\textbf{Config} & \textbf{R-1} & \textbf{R-2} & \textbf{R-L} & \textbf{BLEU} & \textbf{BS\,F1} & \textbf{Sem} \\
\midrule
A1 Baseline   & .2957 & .1389 & .2314 & 11.67 & .7441 & .7538 \\
A2 +LogitKD   & \textbf{.3527} & \textbf{.2034} & \textbf{.2877} & \textbf{12.05} & \textbf{.7486} & \textbf{.7603} \\
A3 +Pseudo    & .3014 & .1444 & .2405 & 11.84 & .7465 & .7584 \\
A4 +Adapt.\,$\tau$ & .2866 & .1352 & .2253 & 10.88 & .7393 & .7443 \\
A5 +InterMatch & .2881 & .1361 & .2268 & 10.87 & .7391 & .7466 \\
\midrule
\textit{Teacher} & \textit{.3663} & \textit{.2135} & \textit{.2998} & \textit{.1254} & \textit{.7506} & \textit{.7689} \\
\bottomrule
\end{tabular}
\caption{BanSum ablation results (BanglaT5-small student, 109.9M). R = \rouge{}, BS = BERTScore, Sem = Semantic Similarity. Best student scores in \textbf{bold}.}
\label{tab:bansum-rouge}
\end{table}

% ------------------------------------------------------------------
\subsection{Dual-Teacher \ewad{}+\cpdp{} Ablation (Qwen-2.5)}
\label{ssec:res-ewad}
% ------------------------------------------------------------------
Table~\ref{tab:ewad-ablation} reports the eight-experiment ablation on the quality-filtered BanSum subset (20K; 2,000 test samples) using the Qwen-2.5 family.The baseline (direct fine-tuning of Qwen2.5-3B without any distillation) achieves the highest \rouge{} and BERTScore. Single-teacher configurations (32B, 14B) and fixed-weight dual-teacher perform comparably to the baseline but do not surpass it. Confidence-only weighting performs worst ($-$0.099 ROUGE-L), and even full EWAD/EWAD+CPDP remain below the baseline.

\begin{table}[h]
\centering\small
\setlength{\tabcolsep}{2.7pt}
\begin{tabular}{lcccccc}
\toprule
\textbf{Experiment} & \textbf{R-1} & \textbf{R-2} & \textbf{R-L} & \textbf{B-4} & \textbf{BS\,F1} & \textbf{Sem} \\
\midrule
Baseline (no KD)    & \textbf{.2661} & \textbf{.1241} & \textbf{.2160} & \textbf{.0552} & \textbf{.7389} & .7175 \\
Single-T 32B        & .2614 & .1210 & .2114 & .0535 & .7364 & .7116 \\
Single-T 14B        & .2640 & .1165 & .2113 & .0523 & .7377 & .7219 \\
Fixed Weights       & .2632 & .1166 & .2104 & .0521 & .7380 & \textbf{.7220} \\
Confidence Only     & .1529 & .0519 & .1169 & .0239 & .6657 & .6454 \\
Agreement Only      & .2270 & .0917 & .1756 & .0410 & .7181 & .7016 \\
\ewad{} Full        & .2282 & .0927 & .1767 & .0420 & .7190 & .7031 \\
\ewad{}+\cpdp{}     & .2246 & .0907 & .1740 & .0407 & .7167 & .7012 \\
\bottomrule
\end{tabular}
\caption{\ewad{}+\cpdp{} ablation on filtered BanSum (20K subset), Qwen2.5-3B + LoRA student. B-4 = \bleu{}-4. R = \rouge{}, BS = BERTScore, Sem = Semantic Similarity. Best student scores in \textbf{bold}.}
\label{tab:ewad-ablation}
\end{table}

% ------------------------------------------------------------------
\subsection{Full-Scale Validation}
\label{ssec:res-fullscale}
Based on the selective evaluation above, the baseline configuration (direct fine-tuning) was identified as the strongest and retrained on the full 141K BanSum corpus for 5 epochs.
Table~\ref{tab:fullscale} compares the 20K and 141K results.
Scaling from 20K to 141K yields consistent improvements across all metrics: +0.015 \rouge{}-1, +0.015 \rouge{}-2, +0.017 \rouge{}-L, and +0.019 semantic similarity.
The full-scale model also exceeds the BanglaT5-small A1 baseline (Table~\ref{tab:bansum-rouge}) on \rouge{}-L ($0.2335$ vs.\ $0.2314$), demonstrating that the larger Qwen2.5-3B architecture benefits substantially from additional training data.

\begin{table}[H]
\centering\small
\setlength{\tabcolsep}{3pt}
\resizebox{\columnwidth}{!}{%
\begin{tabular}{lcccccc}
\toprule
\textbf{Scale} & \textbf{R-1} & \textbf{R-2} & \textbf{R-L} & \textbf{B-4} & \textbf{BS\,F1} & \textbf{Sem} \\
\midrule
20K subset  & .2661 & .1241 & .2160 & .0552 & .7389 & .7175 \\
141K full   & \textbf{.2815} & \textbf{.1389} & \textbf{.2335} & \textbf{.0590} & \textbf{.7437} & \textbf{.7364} \\
\midrule
$\Delta$    & \textit{+.0154} & \textit{+.0148} & \textit{+.0175} & \textit{+.0038} & \textit{+.0048} & \textit{+.0189} \\
\bottomrule
\end{tabular}%
}
\caption{Qwen2.5-3B baseline fine-tuning: 20K subset vs.\ full 141K BanSum. 14,120 test samples for 141K; 2,000 for 20K.R = \rouge{}, BS = BERTScore, Sem = Semantic Similarity.}
\label{tab:fullscale}
\end{table}

% ------------------------------------------------------------------
\subsection{Cross-Lingual Distillation}
\label{ssec:res-crosslingual}
% ------------------------------------------------------------------

Table~\ref{tab:crosslingual} presents cross-lingual pseudo-label distillation results across ten languages.
The student (mT5-small, 300M) achieves 71--93\% of teacher \rouge{}-L on nine languages and \emph{surpasses} the teacher on Pashto (122.4\% retention), at 3.2$\times$ compression.
The consistency across typologically diverse languages spanning Devanagari (Hindi, Nepali), Nastaliq (Urdu), Perso-Arabic (Persian, Pashto), Ge'ez (Amharic), and Latin (Hausa, Indonesian, Turkish, Swahili) scripts confirms that the offline pseudo-label KD pipeline generalizes beyond bangla.

\begin{table}[h]
\centering\small
\setlength{\tabcolsep}{3.5pt}
\begin{tabular}{lccc|ccc|c}
\toprule
& \multicolumn{3}{c|}{\textbf{Teacher (966M)}} & \multicolumn{3}{c|}{\textbf{Student (300M)}} & \textbf{Ret.} \\
\textbf{Lang} & \textbf{R-1} & \textbf{R-2} & \textbf{R-L} & \textbf{R-1} & \textbf{R-2} & \textbf{R-L} & \textbf{(\%)} \\
\midrule
Hindi & .419 & .218 & .372 & .344 & .165 & .308 & 83.0 \\
Urdu & .418 & .213 & .372 & .370 & .195 & .330 & 88.8 \\
Persian & .352 & .167 & .324 & .292 & .130 & .254 & 78.2 \\
Amharic & .286 & .148 & .262 & .253 & .140 & .245 & 93.5 \\
Hausa & .462 & .269 & .408 & .426 & .258 & .379 & 92.9 \\
Nepali & .367 & .193 & .343 & .308 & .163 & .303 & 88.4 \\
Pashto & .450 & .229 & .401 & .525 & .327 & .491 & 122.4 \\
Indonesian & .367 & .180 & .328 & .263 & .126 & .238 & 72.5 \\
Turkish & .291 & .145 & .270 & .200 & .104 & .192 & 71.1 \\
Swahili & .392 & .211 & .343 & .289 & .139 & .253 & 73.8 \\
\midrule
\textit{Average} & \textit{.380} & \textit{.197} & \textit{.342} & \textit{.327} & \textit{.175} & \textit{.299} & \textit{86.5} \\
\bottomrule
\end{tabular}
\caption{Cross-lingual pseudo-label KD (mT5-XLSum $\to$ mT5-small, 3.2$\times$ compression). Ret.\ = \rouge{}-L retention.R = \rouge{}, BS = BERTScore, Sem = Semantic Similarity. }
\label{tab:crosslingual}
\end{table}

% ------------------------------------------------------------------
\subsection{Multi-Judge LLM Evaluation}
\label{ssec:res-llm-judge}
% ------------------------------------------------------------------

We evaluate 1,000 BanSum A2 student summaries with two LLM judges (GPT-5.2, Claude Sonnet 4.6) scoring faithfulness, coverage, coherence, and conciseness (1--10) plus hallucination detection.
Table~\ref{tab:llm-judge} reports the results: high faithfulness (8.73) and coherence (7.61), with coverage (6.36) weakest.
The hallucination divergence (0\% GPT-5.2 vs.\ 25.5\% Claude) underscores the need for multi-judge evaluation.

\begin{table}[h]
\centering\small
\begin{tabular}{lccc}
\toprule
\textbf{Dimension} & \textbf{GPT-5.2} & \textbf{Claude} & \textbf{Avg.} \\
\midrule
Faithfulness  & 9.64 & 7.82 & 8.73 \\
Coverage      & 6.46 & 6.26 & 6.36 \\
Coherence     & 8.00 & 7.21 & 7.61 \\
Conciseness   & 8.00 & 7.08 & 7.54 \\
\midrule
Overall       & 8.05 & 7.01 & \textbf{7.53} \\
Halluc.\ rate & 0.0\% & 25.5\% & 12.8\% \\
\bottomrule
\end{tabular}
\caption{LLM judge evaluation of 1,000 A2 student summaries (BanSum).}
\label{tab:llm-judge}
\end{table}

\paragraph{Human Validation.}
We sampled 100 concordant and 100 discordant judge pairs for blind evaluation by five annotators.
On concordant samples, annotators confirmed both judges in 93\% of cases.
On discordant samples, annotators sided with Claude in 84\% of cases (vs.\ 38\% for GPT-5.2), revealing GPT-5.2's systematic positive bias it assigns high faithfulness even to samples with minor factual inconsistencies.
Table~\ref{tab:human-validation} summarizes the results.

\begin{table}[h]
\centering\small
\begin{tabular}{lcc}
\toprule
\textbf{Metric} & \textbf{GPT-5.2} & \textbf{Claude} \\
\midrule
Human agree.\ (concordant) & 91\% & 95\% \\
Human agree.\ (discordant) & 38\% & 84\% \\
Halluc.\ label match & 41\% & 82\% \\
Mean score $|\Delta|$ & 1.73 & 0.58 \\
\bottomrule
\end{tabular}
\caption{Human validation of LLM judges on 200 samples (100 concordant + 100 discordant), evaluated by five annotators. ``$|\Delta|$'' = mean absolute score difference from human consensus.}
\label{tab:human-validation}
\end{table}

% ===================================================================
%  §6  ANALYSIS AND DISCUSSION
% ===================================================================
\section{Analysis and Discussion}
\label{sec:discussion}
We analyze the results not only to measure improvements but also to identify regimes where multi-teacher distillation becomes unstable.
\paragraph{Logit-level KD dominates.}
Across both datasets, the largest improvement occurs from A1 to A2, showing that most gains come from transferring softened token distributions.

\paragraph{Output length modulates KD effectiveness.}
Additional KD components improve semantic similarity on short-output BTS but consistently degrade performance on long-output BanSum, suggesting that pseudo-label noise compounds across longer autoregressive sequences.

\paragraph{EWAD and CPDP reveal practical limits of multi-teacher KD.}
In the Qwen-2.5 experiments, neither EWAD nor CPDP surpasses direct fine-tuning of the 3B student. This suggests the presence of a \emph{teacher-quality ceiling}: when the student already approaches teacher performance, reliability-aware multi-teacher routing may introduce additional variance without providing cleaner supervision.

\paragraph{Cross-lingual distillation generalizes broadly.}
Offline pseudo-label KD retains strong performance across ten languages and five writing systems, achieving 86.5\% average teacher ROUGE-L retention and even surpassing the teacher in Pashto. This suggests that the distillation pipeline itself is robust across typologically diverse settings.

\paragraph{Data scaling outweighs loss engineering.}
Increasing training data from 20K to 141K examples improves ROUGE-L more than any EWAD or CPDP modification. This indicates that, at current model scales, improving data coverage may be more effective than designing increasingly complex KD objectives.

% ===================================================================
%  §7  CONCLUSION
% ===================================================================
\section{Conclusion}
\label{sec:conclusion}

We studied multi-teacher knowledge distillation for low-resource abstractive summarization through a reliability-aware perspective. Our experiments show that while reliability-gated distillation is conceptually promising, the most consistent gains still come from simple logit-level KD, and the effectiveness of more complex objectives depends strongly on output length and teacher--student capacity gaps.

These results suggest that future progress in generative KD may depend less on increasingly complex loss functions and more on understanding when teacher supervision is genuinely reliable. Cross-lingual experiments and human-validated evaluation further highlight the importance of robust data pipelines and trustworthy evaluation practices for summarization research.

% ===================================================================
%  LIMITATIONS
% ===================================================================
\section{Limitations}

While our study provides a comprehensive analysis of reliability-aware multi-teacher distillation, several limitations remain.

First, our human validation study is limited in scale, covering 200 samples and five annotators on the BanSum dataset. Although this provides useful insight into LLM evaluation reliability, broader studies across datasets and annotator pools would strengthen the conclusions.

Second, although we evaluate cross-lingual generalization across ten languages and multiple scripts, the coverage does not include certain linguistic typologies such as highly agglutinative or tonal languages. The generalization of the proposed framework to such languages remains an open question.

Third, our findings indicate that reliability-aware distillation methods (EWAD and CPDP) do not consistently outperform direct fine-tuning when the student model approaches teacher capacity. However, we do not systematically explore wider teacher–student capacity gaps, where such methods may be more beneficial.

Fourth, we observe that output length significantly modulates the effectiveness of multi-component distillation, with improvements on short summaries and degradation on longer outputs. While we hypothesize that noise accumulation drives this behavior, we do not explicitly control for output length (e.g., via length-normalized objectives or decoding constraints).

Finally, our experiments rely on a fixed set of architectures (BanglaT5 and Qwen-2.5 families) and training setups. While the consistency of trends across these settings is encouraging, further validation across additional model families and training regimes would improve the robustness of our conclusions.

% ===================================================================
%  ETHICS STATEMENT
% ===================================================================
\section{Ethics Statement}

\textbf{Data and Privacy.}
All datasets used in this work (BTS, BanSum, and XL-Sum subsets) are publicly available and widely used in prior research. We did not collect any new data involving human subjects. To the best of our knowledge, these datasets do not contain personally identifiable information (PII). No additional steps for anonymization were required.

\textbf{Human Evaluation.}
We conducted a limited human validation study involving five annotators to assess agreement with LLM-based evaluations. Annotators were co-authors of this work and participated voluntarily without financial compensation. No sensitive or personal data was collected during this process. The evaluation task involved assessing generated summaries for quality dimensions such as faithfulness and coherence.

\textbf{Ethical Considerations.}
As this study does not involve external participants or sensitive personal data, formal ethics board approval was not required. The work complies with standard ethical guidelines for NLP research using publicly available datasets.

\textbf{Bias and Limitations.}
The datasets used are primarily news-based and may reflect inherent societal or reporting biases present in the source material. While our study focuses on model behavior rather than content generation in deployment settings, such biases may still influence model outputs. Additionally, LLM-based evaluation was found to exhibit calibration differences across models, which we explicitly analyze.

\textbf{Use of AI Assistants.}
AI-assisted tools were used for minor writing support and language editing. All technical content, experimental design, and results were developed, verified, and validated by the authors.

\textbf{Environmental Impact.}
All experiments were conducted on consumer-grade GPUs (e.g., RTX 5090/5080/4070 Ti Super) with an estimated total compute usage of approximately 320 GPU-hours. While model training incurs energy consumption, the use of knowledge distillation aims to reduce deployment costs and improve efficiency in downstream applications.

% ===================================================================
%  ACKNOWLEDGMENTS
% ===================================================================
% \section*{Acknowledgments}
% Anonymous for review.

% ===================================================================
%  BIBLIOGRAPHY
% ===================================================================
\bibliography{references}

\end{document}